\documentclass[sigconf]{acmart}

\usepackage{amsmath,amsfonts}
\usepackage{algorithmic}
\usepackage{algorithm}
\usepackage{array}
\usepackage[caption=false,font=normalsize,labelfont=sf,textfont=sf]{subfig}
\usepackage{textcomp}
\usepackage{stfloats}
\usepackage{url}
\usepackage{verbatim}
\usepackage{graphicx}

\usepackage{tcolorbox}
\usepackage{minted}
\usepackage{longtable}
\usepackage{soul} 
\usepackage{color, xcolor} %
\tcbuselibrary{breakable}
\usepackage{listings} 
\usepackage{tabularx}
\usepackage{multirow} 
\usepackage{multicol} 
\usepackage{ulem}
\usepackage{hyperref}
\usepackage[table]{xcolor} 
\usepackage{booktabs}
\usepackage{orcidlink}
\usepackage{booktabs} 
\usepackage{graphicx} 

\AtBeginDocument{%
  }

\setcopyright{acmlicensed}
\copyrightyear{2018}
\acmYear{2018}
\acmDOI{XXXXXXX.XXXXXXX}
\acmConference[Conference acronym 'XX]{Make sure to enter the correct
  conference title from your rights confirmation email}{June 03--05,
  2018}{Woodstock, NY}
\acmISBN{978-1-4503-XXXX-X/2018/06}

\begin{document}

\title{Causal Agent based on Large Language Model}


\author{Kairong Han}
\orcid{0000-0003-0003-2312}
\affiliation{%
  \institution{College of Computer Science and Technology, Zhejiang University}
  \city{Hangzhou}
  \postcode{310027}
  \country{China}}
\email{zju\_handso@163.com}

\author{Kun Kuang}
\authornote{Corresponding author.}
\orcid{0000-0001-7024-9790}
\affiliation{%
  \institution{College of Computer Science and Technology, Zhejiang University}
  \city{Hangzhou}
  \postcode{310027}
  \country{China}}
\email{kunkuang@zju.edu.cn}

\author{Ziyu Zhao}
\orcid{0000-0003-1460-2777}
\affiliation{%
  \institution{College of Computer Science and Technology, Zhejiang University}
  \city{Hangzhou}
  \postcode{310027}
  \country{China}}
\email{benzhao.styx@gmail.com}

\author{Junjian Ye}
\orcid{0009-0001-4614-4662}
\affiliation{%
  \institution{Huawei Technologies Company}
  \country{China}}
\email{yejunjian@huawei.com}

\author{Fei Wu}
\orcid{0000-0003-2139-8807}
\affiliation{%
  \institution{College of Computer Science and Technology, Zhejiang University}
  \city{Hangzhou}
  \postcode{310027}
  \country{China}}
\email{wufei@zju.edu.cn}

\renewcommand{\shortauthors}{Trovato et al.}

\begin{abstract}
The large language model (LLM) has achieved significant success across various domains. However, the inherent complexity of causal problems and causal theory poses challenges in accurately describing them in natural language, making it difficult for LLM to comprehend and use them effectively. Causal methods are not easily conveyed through natural language, which hinders LLM's ability to apply them accurately. Additionally, causal datasets are typically tabular, while LLM excels in handling natural language data, creating a structural mismatch that impedes effective reasoning with tabular data. This lack of causal reasoning capability limits the development of LLM.
To address these challenges, we have equipped the LLM with causal tools within an agent framework, named the Causal Agent, enabling it to tackle causal problems. The causal agent comprises tools, memory, and reasoning modules. In the tool module, the causal agent calls Python code and uses the encapsulated causal function module to align tabular data with natural language. In the reasoning module, the causal agent performs reasoning through multiple iterations with the tools. In the memory module, the causal agent maintains a dictionary instance where the keys are unique names and the values are causal graphs.
To verify the causal ability of the causal agent, we established a \underline{\textbf{Causal}} \underline{\textbf{T}}abular \underline{\textbf{Q}}uestion  \underline{\textbf{A}}nswer (CausalTQA) benchmark consisting of four levels of causal problems: variable level, edge level, causal graph level, and causal effect level. CausalTQA consists of about 1.4K for these four levels questions. Causal agent demonstrates remarkable efficacy on the four-level causal problems, with accuracy rates all above 80\%. Through verification on the real-world dataset QRData, the causal agent is 6\% higher than the original SOTA, demonstrating its strong generalization ability in real-world scenarios. For further insights and implementation details, our code is accessible via the GitHub repository \href{https://github.com/Kairong-Han/Causal_Agent}{here}.
\end{abstract}

\ccsdesc[500]{Computing methodologies~Natural language processing}
\ccsdesc[500]{Computing methodologies~Intelligent agents}


\maketitle

\section{INTRODUCTION}

In recent years, generative artificial intelligence technology has gained significant success, making remarkable achievements in the natural language processing field\cite{kocon2023chatgpt,10.1145/3829078}, image, audio synthesis, etc\cite{9599397,10731578}.         This advancement lays the foundation for propelling research in general artificial intelligence\cite{10711229}, both in terms of framework development and practical implementation.       However, due to the complexity of causal problems, the causal reasoning capabilities of the large language model (LLM) remain insufficient. Causal theory is difficult to describe in natural language that the LLM can understand accurately. Researchers have evaluated the pure causal reasoning abilities of the LLM and found that their pure causal reasoning is close to random\cite{jin2023large}. Additionally, researchers believe that the current LLM is merely "causal parrots" that mimic without truly possessing causal understanding\cite{zečević2023causal}. This inherent limitation severely hampers the performance of large models in tasks requiring causal reasoning.
Moreover, causal datasets are typically tabular data, while large models excel in handling natural language data. Although methods such as TableGPT\cite{sun2024survey,li2023table} have been proposed for processing tabular data, they primarily focus on simplified table manipulations (e.g., copying or swapping rows). When we need to draw causal conclusions based on the analysis of tabular data, the LLM which is not specifically designed cannot directly utilize tabular data and perform reasoning. This structural heterogeneity hinders LLM from effectively reasoning with tabular data. These two limitations restrict the ability of LLM to solve causal problems effectively.
\begin{figure*}[h]
    \centering
    \includegraphics[width=1\linewidth]{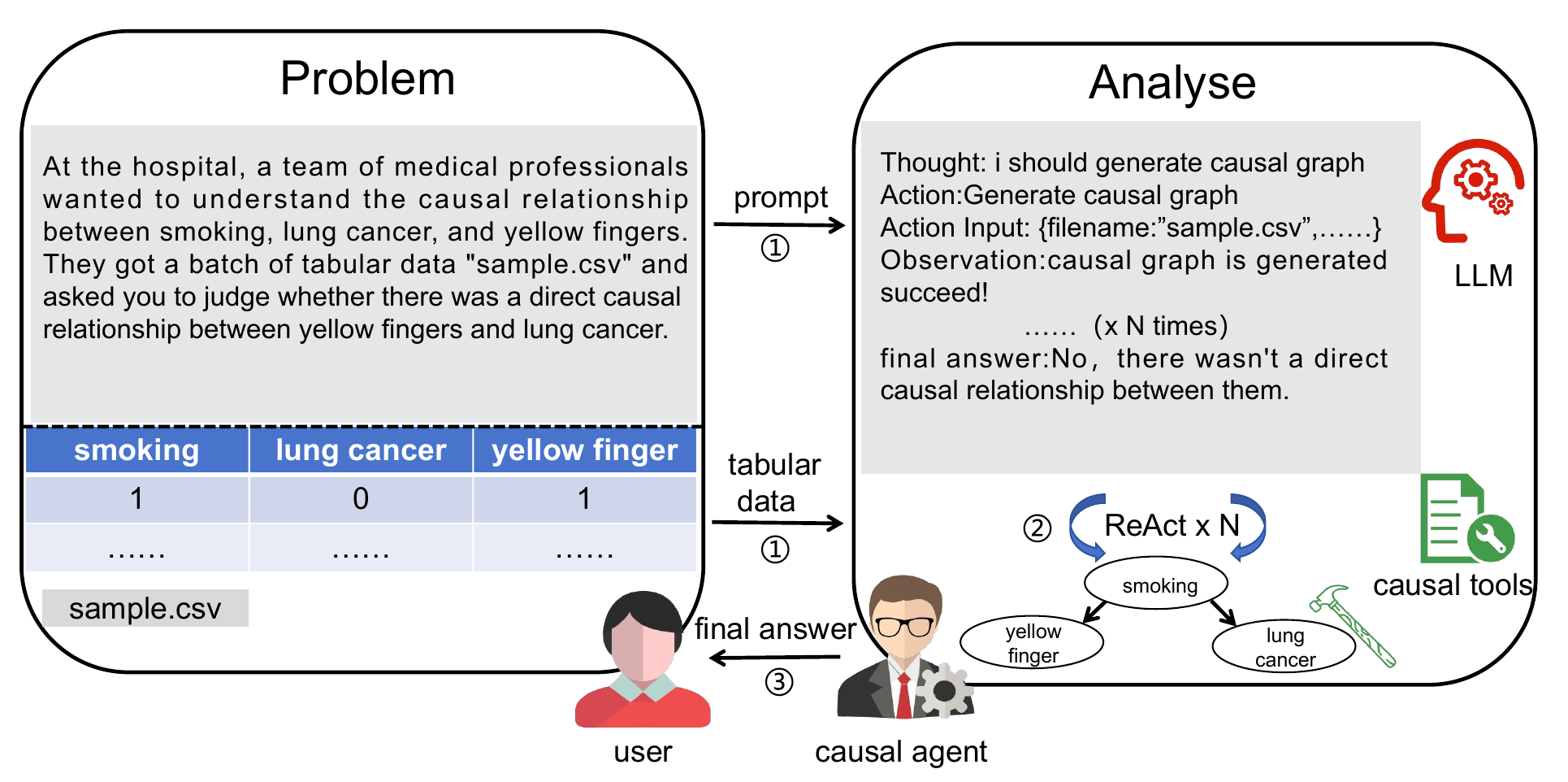}
    \caption{Working flowchart of the causal agent. In the first step, the user inputs a pair of tabular data and the causal problem; In the second step, the causal agent invokes the causal tools (tools module) and uses the ReAct framework (plan module) to conduct multiple rounds of analysis for the tabular data, in which causal agent maintains a dictionary of causal graph names and their instantiations as memory (memory module);  In the third step, the causal agent is combined with the analysis process to produce corresponding answers for the user's problems.}
    \label{fig:示意图}
\end{figure*}

To solve two difficult problems in improving the causal ability of LLM, we propose a causal problem modeling approach from the perspective of the LLM and propose a causal agent framework by guiding LLM to invoke causal tools. We model the causal problems into four levels: variable level, edge level, causal graph level, and causal effect level. The variable level focuses on the agent's judgment and understanding of correlations, the edge level focuses on the agent's examination of causal relationships between variables, the causal graph level focuses on the agent's ability to generate causal graphs, and the causal effect level focuses on the agent's estimation of causal effects between variables for quantitative expression. 

Based on the abovementioned causal problems, we construct a causal agent based on LLM. The causal agent uses causal graphs as an intermediary for causal reasoning in the domain. By leveraging causal discovery algorithms to generate causal graphs and conducting reasoning on these graphs, it bridges the complexity of downstream causal tasks with the challenges posed by tabular data question answering. Specifically, the causal agent is composed of tools, memory, and plan modules, as shown in Figure~\ref{fig:示意图}. In the tools module, the causal agent invokes the causal analysis library in Python programming tools, such as causal-learn\cite{causallearn} and EconML\cite{econml}.  So the causal agent can receive a pair of tabular data and a causal problem description of the data as input.  By invoking causal analysis tools, the tool processes the tabular data and generates natural language conclusions that the causal agent can understand. In the plan module, the causal agent utilizes its text comprehension and reasoning abilities to obtain answers to causal problems in many iterations. In the memory module, the causal agent may need multiple tools to solve a problem. To preserve intermediate results during the planning process, the agent maintains an instantiated dictionary where the keys are names and the values are causal graphs. This special method allows the agent to retrieve the necessary causal graph using the key.
On the one hand, the content of the memory is expressed in a more informative way. On the other hand, using a data structure rather than text as memory can effectively simplify the complexity of prompt design during the reasoning process. 

To verify the causal ability of the causal agent, we established a \underline{\textbf{Causal}} \underline{\textbf{T}}abular \underline{\textbf{Q}}uestion  \underline{\textbf{A}}nswer (CausalTQA) benchmark consisting of four levels of causal problems: variable level, edge level, causal graph level, and causal effect level. CausalTQA consists of about 1.4K for these four levels questions. In a controlled synthetic data setting, the causal agent demonstrates a high accuracy in correctly invoking tools and producing the expected answers at four level questions, with accuracy rates of over 95\% in all three sub-problems for determining correlation at the variable level, over 89\% in all three sub-problems at the edge level, over 81\% in the causal graph level, and 98\% in the causal effect estimation level. Through verification on the real-world dataset QRData, the causal agent is 6\% higher than the original SOTA, demonstrating its strong
generalization ability in real-world scenarios.  

Additionally, our design has the advantage of flexibility and reliability: it allows the causal agent to integrate and call different built-in algorithms, thereby bridging the causal inference and LLM communities. This approach ensures controllability in the model's reasoning process while enabling seamless adaptation to advancements in both fields. The detailed intermediate reasoning process makes the model's decision-making transparent and reliable. 
Our contributions are summarized as follows:
\begin{itemize}
\item	A hierarchical modeling perspective has been proposed for LLM to solve causal problems. This is a new setting, and the problem is set to be data-driven, where the LLM answers causal questions about tabular data when users input a pair of tabular data and causal questions. We focus on four-level questions for causal agents to solve causal problems, denoted as variable level, edge level, causal graph level, and causal effect level. We propose the CausalTQA benchmark, approximately 1.4K in size for the four levels of problems, covering nine sub-problems in total at four levels;
\item	The causal agent has been proposed to empower LLM with the ability to solve causal problems.  In this framework, we use LLM to invoke causal tools and iterate many times to analyze and solve causal problems.  Thus, heterogeneous data alignment is achieved between natural language input for large models and tabular data input for causal problems. In addition, the causal graph is used as the core of the interaction between different tools to improve the reasoning ability of the causal agent. The causal agent framework that empowers causal reasoning through the use of causal tools has good Interpretability and reliability;

\item	In the CausalTQA benchmark, the causal agent achieved high accuracy in the four levels of causal problems. Specifically, all three sub-problems at the variable level achieved an accuracy of over 95\%, all three sub-problems at the edge level achieved an accuracy of over 89\%, all three sub-problems at the causal graph level achieved an accuracy of over 81\%, and all two sub-problems at the causal effect level achieved an accuracy of over 98\%. Through verification on the real dataset QRData\cite{liu2024llmscapabledatabasedstatistical}, the causal agent is 6\% higher than the original SOTA, demonstrating its strong generalization ability in real scenarios;
\item It demonstrates the potential of the causal agent for automated causal reasoning on tabular data, facilitates the use of causal tools, and positively impacts the widespread popularization of causal theory;
\end{itemize}

\section{MATERIALS AND METHODS}

\subsection{Modeling Causal Problems from the Perspective of LLM}\label{Modeling causal problems from the perspective of LLM}

Despite the development of LLM, like GPT, demonstrating strong natural language understanding and question-answering capabilities, there is still a significant gap between the problem paradigms that data-driven causal focuses on tabular data but LLM focuses on the field of natural language processing. Furthermore, LLM struggles to truly understand and handle the intricate causal relationships inherent in complex data. 

Therefore, it is meaningful to re-establish a causal problem framework from the perspective of the LLM. This has a significant impact on evaluating the causal ability of the LLM and enhancing its causal ability. To model causal problems within the field of LLM, we formulate our settings as follows:

Let $T \in R^{n\times c}$be a tabular data with $n$ rows and $c$ columns, in which each row $t_i$ is a piece of data, and each column $c_i$ represents a variable. So 
$$T=\{t_i\}_{i=0}^{n}$$

We formalize the causal problem space as a $Q$ and $q_i \in Q$ is one question in the form of natural language. We combine the tabular data and the problem description by Cartesian product to create the dataset $D$ and each item $ d_i \in R^{n\times c} \times Q$. So

$$D = \{d_i\} = \{(T_i,q_i)\in R^{n\times c} \times Q \}$$

The user inputs a pair of $(T_i,q_i)$ samples from $D$, and then the causal agent analyses the tabular data $T_i$ and the causal problem $Q$ to generate a reasonable answer $A$. The format of answer $A$  is not limited to the form of natural language. $A$ can also be a causal graph or other non-textual data to explain the question clearly.



Due to the complex diversity of causal problems, we simplify the problem and conduct the necessary modeling.  We categorize the causal problems into four major levels based on the differences in problem granularity and objects: variable level, edge level, causal graph level, and causal effect level.  The variable level corresponds to the first level of the causal ladder, correlation, aiming to endow LLM with the ability to infer and understand correlations between variables in tabular data.  The edge level builds beyond correlation, aiming to endow LLM with the ability to understand and analyze causal edge relationships between variables.  The causal graph level shifts the perspective to a more macroscopic dimension, examining the LLM's capabilities of generating causal graphs.  The causal effect level aims to endow LLM with the ability to quantify the causal effects between variables. We will discuss four levels of modeling methods in detail below.

\subsubsection{Variable level}

At the variable level, we focus on determining the correlation between different variables, which is the first level of the causal ladder.    To obtain correlation from tabular data, we transform the problem of correlation testing into independence testing.    That is given variables $V_i$ and $V_j$, determining whether they are independent or conditional independent under variables $\{V_k\}_{k=1}^N$.    If two variables are correlated, they are statistically dependent, and vice versa.    Through such modeling methods, we aim to test the causal agent with the ability to analyze correlations.    Specifically, we divide the problem of correlation into two sub-classes: direct independence testing and conditional independence testing. The difference between them lies in whether condition variables are given when judging independence. In particular, direct independence testing can be regarded as the number of condition variables is zero. To more finely measure the model's capabilities, we further divide conditional independence testing into independence testing under a single condition and independence testing under multiple numbers of conditions, in which the difference is whether the number of conditions is one or beyond one.

\subsubsection{Causal graph level}\label{causal graph level}

At the causal graph level, the focus is more macroscopic, examining whether the causal agent possesses the capability to generate a causal graph. The causal graph is a directed acyclic graph (DAG), and in DAG the direction of edges represents causal relationships.  In this article, we choose the PC algorithm\cite{spirtes2000causation} as the method for generating causal graphs, which generates Markov equivalence classes of causal graphs without considering the presence and influence of unobserved variables.  Modeling the capabilities of intelligence at the level of the causal graph involves two categories:  generating a causal graph that includes all variables in tabular data, and the other generating a partial causal graph that includes only a subset of variables in tabular data.  The capability to generate causal graphs and to reason on these graphs can effectively guide the agent to understand causal relationships and discern true causal connections amidst the fog of spurious correlations.

\subsubsection{Edge level}

At the edge level modeling, we still consider the relationships between variables.   Instead of the associations from a statistical correlation, we focus on the deeper causal relationships between variables from a causal viewpoint.      Unlike quantitative estimation of causal effects, edge-level modeling provides qualitative analysis results that need to reflect the true relationships of the edges in the causal graph reconstructed from tabular data.    We consider the following three types of relationships: direct causal relationship, collider relationship, and confound relationship.  As discussed in Section \ref{causal graph level}, we used the PC algorithm to generate Markov equivalence classes for causal graphs, therefore we  formalize three types of relationships as follows:

Denote $G$ as a Markov equivalence class generated by the PC algorithm from tabular data, containing edges set $ \{ <V_i,V_j>\}$. There are two types of edges: directed edges and undirected edges, so $<V_i,V_j> \in \{ \xrightarrow {},- \}$. 

We denote direct causal relationships as $V_i$ directly causes $V_j$, reflected in the causal graph $G$ as existence edge $V_i\xrightarrow {}V_j $. We denote the collider relationship as $V_i$ and $V_j$ directly cause a common variable $V_k$ , reflected in the causal graph as existence $V_i\xrightarrow {}V_k $ and $V_j\xrightarrow {}V_k $ . We denote the confounding relationship as the presence of unobstructed backdoor paths between $V_i$ and $V_j$, reflected in the causal graph as  $V_i\xleftarrow{} ... \xrightarrow {}V_j $

\subsubsection{Causal effect level}

The causal effect level attempts to quantify how the outcome for an individual or system would differ if it experienced a certain intervention. 
We expect the causal agent not only to utilize causal explanations for qualitative analysis but also to employ classical causal inference for quantitative interpretation.  To simplify the problem, at the level of causal effects, we only consider the quantitative calculation of the ATE, denoted as $E(Y(T=t_1)-Y(T=t_0))$, from tabular data.
Modeling at the granularity of causal effects can equip the causal agent with a more fine-grained causal perception capability.

\subsection{Causal Agent Framework Based on LLM }

Causal agent framework consists of three modules: tools module, memory module, and plan module.      In terms of tools, to align the tabular data with natural language, we invoke causal analysis tools that can accept tabular data as input.     For the output of tools, we use prompts to interpret and explain the results, enabling the causal agent to understand the output.   In the planning aspect, inspired by the ReAct framework\cite{yao2023react}, through multiple rounds of reflection, we continuously invoke causal analysis tools and reflect on their output, considering whether we can derive the answer to the original question based on the agent's understanding of the causal question. If the answer to the question cannot be derived, we continue to iterate and reflect until we reach the final answer or limited iteration times.      Besides, to better understand tools' usage, we use in-context learning and one-shot examples to empower the causal agent. A manual handwritten example is designed to use all tools to guide the causal agent in invoking and understanding the tool.     In terms of memory, we store the output of the causal analysis tools in a dictionary in memory as short-term memory, ensuring that the agent can continuously access the causal graph before the final answer is obtained.

\subsubsection{Tools}
\begin{figure*}[h]
    \centering
    \includegraphics[width=1\linewidth]{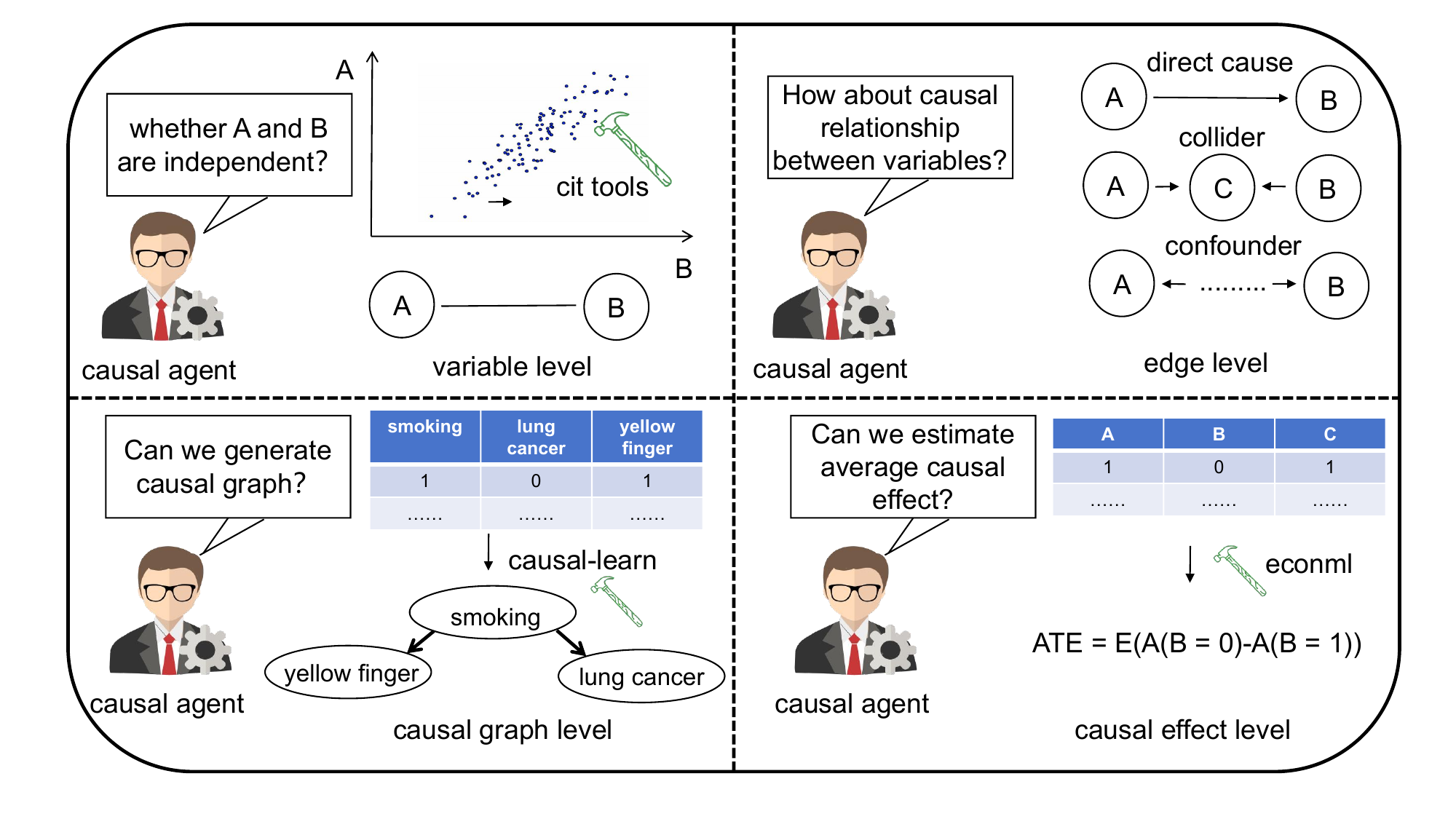}
    \caption{Causal Agent Tools Usage Diagram: Different tools are used to address causal problems at four levels. 
    }
    \label{fig:shiyitu2}
\end{figure*}

\newtcolorbox[auto counter, number within=section]{mybox}[2][]{colframe=blue!30!gray, colback=yellow!10, 
coltitle=black, fonttitle=\bfseries, title={\tikz[baseline]{\node[yellow!20, text=black, anchor=base] {\textbf{#2}};}},
sharp corners=southwest, breakable, #1}

As shown in Figure \ref{fig:shiyitu2}, the causal agent invokes causal analysis tools to analyze tabular data, thereby compensating for the LLM's shortcomings in handling tabular data.       This approach aligns tabular data with causal conclusions and enhances the LLM's causal capabilities through tool invocation.       Specifically, our causal analysis tools select the library causal-learn\cite{zheng2024causal} for causal discovery and EconML\cite{econml} for causal inference.       Starting from the perspective of modeling causal problems for the LLM, we have designed specific tool functions at the variable level, edge level, causal graph level, and causal effect level.       To make the tool functions easily invoked by the LLM, we have re-encapsulated the interfaces, changing the tool inputs to JSON string format, and using manual rules and handwritten prompt templates to help the large model better understand the meaning of the tool outputs.  The prompt details can be found in the code repository. An example is shown below:

\begin{mybox}[title={Condition Independent Test}]

Useful for when you need to test the independence or d-separation of variable A and variable B, conditioned on variable C. The input should be a JSON with the format below: 
 \{"filename":...,"interesting\_var":[...],"condition":[...]\}

An interesting\_var is a list of variables the user is interested in. For example, if the user wants to test independence (d-separate) between X and Y conditions on Z, W, Q, the interesting\_var is ["X","Y"], the condition is ["Z","W","Q"]. Condition is $[]$ if no condition is provided.
\end{mybox}

At the variable level, we invoke the conditional independence test class in causal-learn and use Fisherz\cite{fisher1921014} as an independence test method.    At the causal graph level, since there are no unobserved variables in our data assumptions, we invoke the PC algorithm to generate the Markov equivalence class of the causal graph.       It should be noted that when generating a partial causal graph, we still use the PC algorithm.       However, in this case, the variables not included in the arguments are unobserved variables for partial causal graphs. We think that this situation should be controlled by the user rather than the agent actively changing the causal discovery algorithm, such as the FCI\cite{spirtes2013causal} algorithm that can handle unobserved confounders.       This design maintains the reliability of the agent's behavior and facilitates user interaction with the agent.

At the edge level, we use the tool's prompt template to guide the LLM to use the causal graph generation algorithm and obtain the Markov equivalence class of the causal graph, then judge the relationship between the edges. For undirected edges that the PC algorithm can not determine,  the tools will categorically discuss the direction of the edge to conclude.      We focus on three sub-issues at the edge level: direct cause,  confounding, and collider.       For judging the cause relationship, we consider whether there is a directed edge directly connecting the two variables in the output $G$ of the PC algorithm. For judging confounding, we consider whether there exist unblocked backdoor paths between the two nodes. For judging a collider, we only consider the collider "V" structure, such as $V_i\xrightarrow[]{}V_k\xleftarrow[]{}V_j.$

At the level of causal effects, the causal agent invokes the LinearDML algorithm in the EconML library, where the user needs to specify which variables are covariates. The causal agent first uses the LinearDML algorithm to train the data.   During the training stage, we choose the default parameters of the LinearDML algorithm in the EconML library, and the covariates are specified by the user's prompts input.   After the training stage, the tool outputs an estimator of the average causal effect on the same model, using covariates consistent with those used during training.   

\subsubsection{Plan process}

Inspired by the ReAct framework, the causal agent adopts an iterative multi-turn dialogue approach, using prompt templates to facilitate interaction and understanding between the causal agent and the tools. During the process of invoking causal analysis tools, the agent must continuously think and experiment to determine the next action. This process is crucial for the causal agent, as solving complex causal problems may require the use of multiple causal tools more than one, as well as the observation and integration of different outputs from these tools. Therefore, the causal agent needs to repeatedly invoke these tools, observe their output, and engage in deep thinking to arrive at the solution.

Specifically, in each round of the dialogue, the user provides a problem description $q$ and the table data $T$ to be analyzed as input $(q, T)$.   Guided by the prompt, from the first round of each round $i$, the causal agent will generate a thought $O_i^t$ and use a tool as action $a_i$.   The tool will accept $a_i$ and generate observation $O_i^{obs}$.   Based on this observation, the causal agent attempts to answer the original question description.   If it is still unable to answer, it concatenates this round's observation as input to continue the next round.   The input for the next round is $(q,T, O_1^t,a_1,O_1^{obs},O_2^t,a_2,O_2^{obs},\ldots,O_i^{obs})$, until the causal agent comes up with the final answer or reaches the maximum times of iterations. Specific examples are as follows:

\begin{mybox}[title={Plan Process}]

\textcolor{red}{Input prompt}:

Consider four elements: age, weight, sleep time, and cancer \dots whether there is a direct causal effect of weight on cancer. CSV data is stored in data.csv.

\textcolor{red}{Input data}: data.csv
\\
\noindent\rule{\linewidth}{0.4pt}

\textcolor{red}{Thought}: We should start by generating a causal graph to analyze the relationships between these variables.

\textcolor{red}{Action}: Generate Causal

\textcolor{red}{Action Input}: \{"filename": "data.csv", "analyze relationship": "True"\}

\noindent\rule{\linewidth}{0.4pt}

\textcolor{red}{Observation}: Causal graph named 'data' is generated successfully!

\textcolor{red}{Thought}: Now that we have generated the causal graph, we can use it to determine if age is a confounding factor in the relationship between weight and cancer.

\textcolor{red}{Action}: Determine Confounder

\textcolor{red}{Action Input}: \{"cg\_name": "data", "interesting\_var": ["weight", "cancer"]\}

\noindent\rule{\linewidth}{0.4pt}

\textcolor{red}{Observation}: \dots (multi times iterators)

\textcolor{red}{Final Answer}: There is no direct causal effect of weight on cancer.
\end{mybox}

\subsubsection{Memory}

\begin{figure}[h]
    \centering
    \includegraphics[width=1\linewidth]{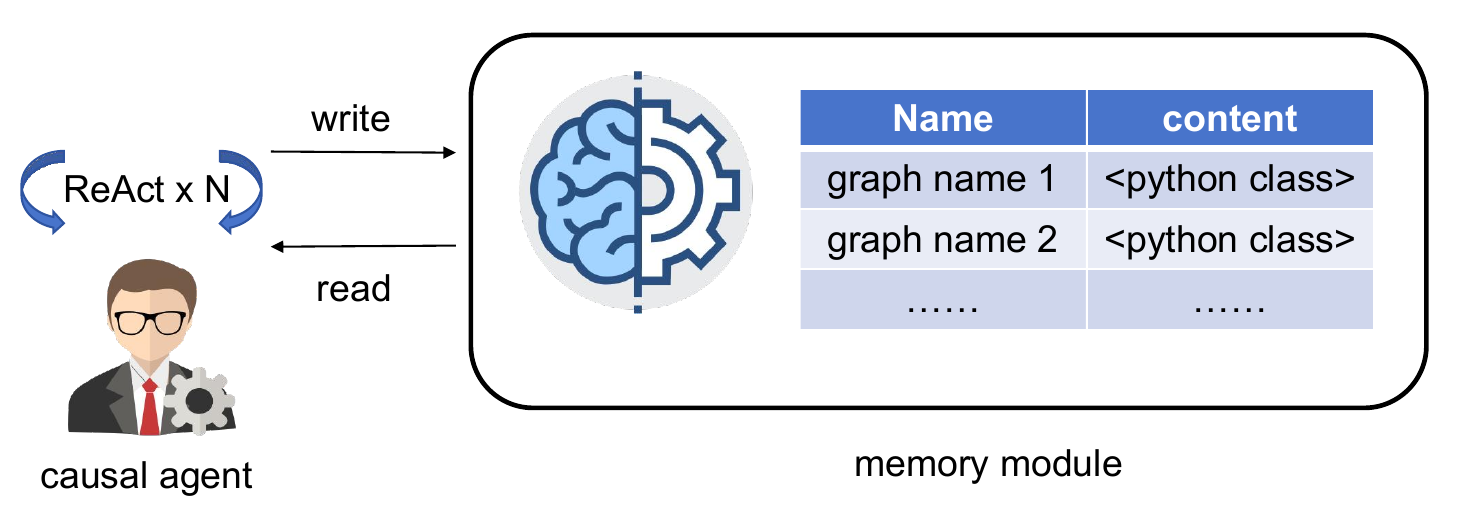}
    \caption{Causal Agent Memory Module Diagram: During the reasoning process, the causal agent maintains a memory index in its memory. The index names are in natural language form, while the index content consists of data structures such as causal graph instances containing richer information. 
    }
    \label{fig:memorymodule}
\end{figure}

During the reasoning and interaction process of the causal agent, it needs to maintain the currently generated causal graph and use this graph in subsequent causal relationship judgments. 
Therefore, in the implementation of the causal agent's memory, the memory is not a textual form of data but the data corresponding to the causal graph Python class instance. The causal agent maintains a dictionary as memory, adding an entry and establishing a name index during memory writing, and using the index to read the corresponding causal graph information during multi-turn dialogues, as shown in Figure \ref{fig:memorymodule}. The memory module of the causal agent differs from traditional LLM-based agent memory in that the stored data structure is not natural language but rather more informative as an abstract expression of memory. By maintaining a causal graph and performing reasoning over it, the memory module acts as an intermediary that connects different causal tools. This enables the tools to form invocation chains, thereby facilitating complex reasoning processes.
 
\section{RESULTS}
To test the causal agent, we conducted synthetic data experiments on CausalTQA and real data QRData verification. The data generation process is shown in Section \ref{DPG}. The experiment on CausalTQA is shown in \ref{Causal Problem Test Result}, and the real data verification on QRData is shown in \ref{QRDATA verify}. We use gpt-3.5-turbo as the default decision-making core of the causal agent, and calculate the cost of the causal agent in \ref{wentikaixiao}. The use of different models as the causal agent decision core, and the comparison with the existing multi-agent method are shown in Section \ref{Discussion on Multi-Agent Systems and Different LLMs}.


\subsection{Data Generate Process}\label{DPG}
\begin{figure*}[h]
    \centering
    \includegraphics[width=1\linewidth]{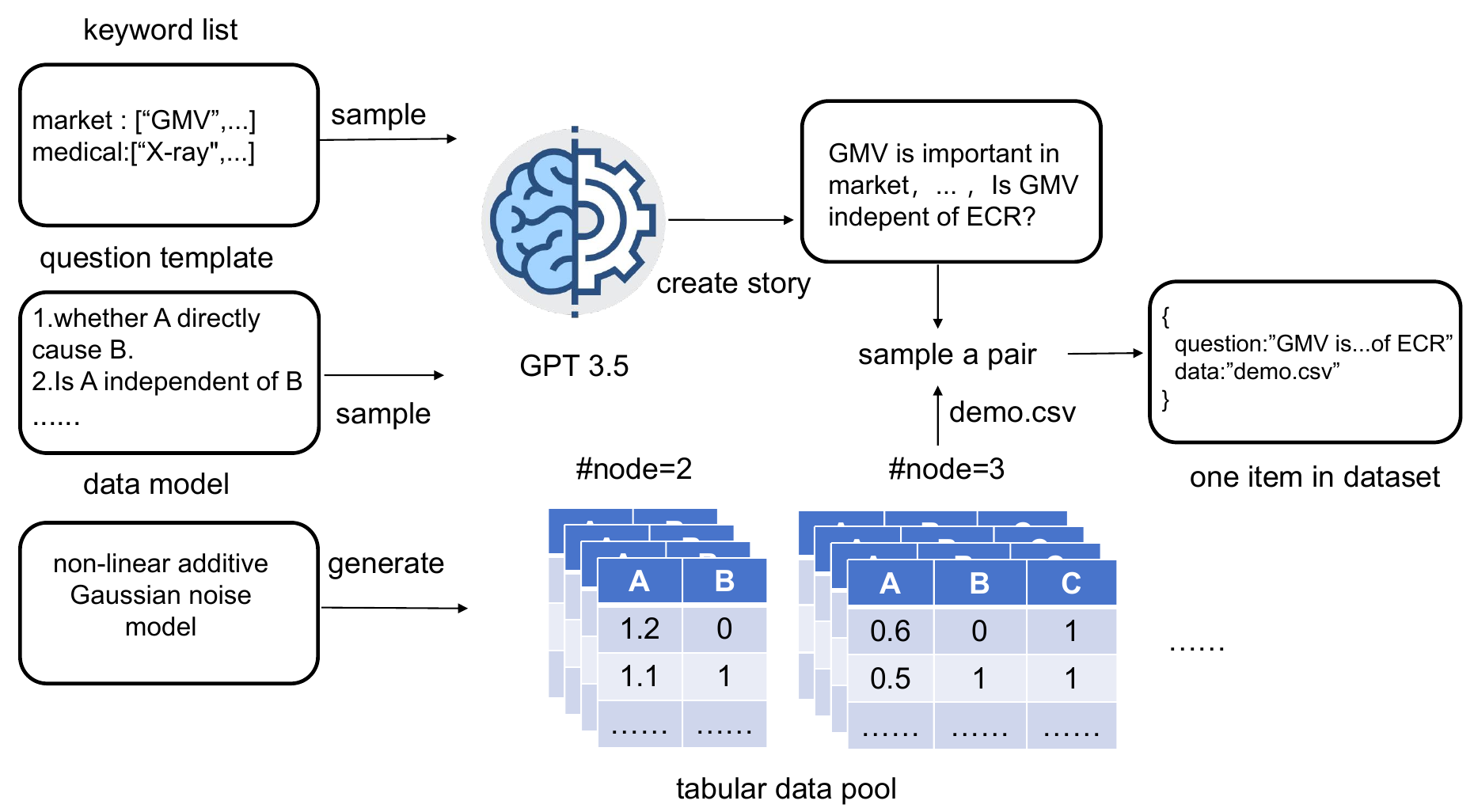}
    \caption{Overview of Data Generation Process: The final generated data consists of a pair of causal questions and tabular data. 
    Initially, GPT-3.5 samples a set of descriptions from keyword and question lists, forming a group of descriptions, and generates detailed narratives and causal question descriptions based on these. Subsequently, a non-linear additive Gaussian model is employed to generate a pool of tabular data. Data is then extracted from this pool and combined with the causal question descriptions to form a single data entry.
    }
    \label{fig:datagenerateprocess}
\end{figure*}
\subsubsection{tabular data}

To generate the tabular data needed for the test dataset, we adopt the method used in the work by \cite{rolland2022scorematchingenablescausal}. Specifically,  our data follows non-linear additive Gaussian noise models, $X \in R^d$ is generated using the following model:
$$X_i = f_i(pa_i(X) )+ \epsilon_i $$
$ i = 1, \dots,d$ , where  $pa_i(X)$ selects the coordinates of  $X$  which are parents of node $i$ in some DAG. And $$ \epsilon_i \sim \mathcal{N}(0, \sigma^2)$$ The functions $f_i$ are assumed to be twice continuously differentiable and non-linear in every component. That is, if we denote the parents  $pa_j (X)$ of  $X_j$  by  $X_{k1}, X_{k2}, \dots, X_{kl}$ , then, for all  $a = 1, \dots, l$ , the function  $f_j(X_{k1}, \dots,X_{k_{a-1}},\cdot,X_{k_{a+1}},\dots,X_{kl} )$  is assumed to be nonlinear for some  $X_{k1}, \dots,X_{k_{a-1}},\cdot,X_{k_{a+1}},\dots,X_{kl} \in R^{l - 1}$ 

Through the aforementioned method, we have generated a series of tabular data with node counts ranging from 3 to 10. For tables with the same number of nodes, we generate a series of tabular data with different edge numbers, which range in $ [0,C_{\#node}^2]$. This simulates different scenarios of sparsity and density of real causal graphs. We use all the generated tabular data as a data pool. When generating specific test samples later, we will randomly take a table from the pool that has the same number of nodes as variables required by the question, to be a quantitative expression of the relationships between variables. 

\subsubsection{causal problem descriptions}

To simulate causal issues in real scenarios, we generate a natural language template $T_q$ for the four-level causal questions modeled in Section \ref{Modeling causal problems from the perspective of LLM}. Then we take the medical field and the market field, two common fields for causal inference, as question sources to generate questions' real-world scenes. We first generate 100 keywords related to medical and market as list $L_q$ . Then, we iteratively traversed through node counts from 3 to 10. For a node count of $i$, we randomly drew $i$ keywords from the keyword list as seeds $$K_1,...,K_i \sim L_q$$ 
Subsequently, we used the question template  $t_q^{i} \sim T_q$ to allow GPT-3.5 to construct a possible real scenario using the seed keywords, thus forming a piece of data.$$ description = GPT(K1,...,K_i,t_q^i)$$ Note that the keywords that come from the list are randomly drawn, so there may be no causal relationship between them. 
This special design makes the causal agent focus on the tabular data during the process of causal analysis, leading to data-driven rather than semantic causal information between variables for causal analysis.

\subsubsection{Statistics of the CausalTQA}

\begin{figure}[h!]
    \centering
    \includegraphics[width=1\linewidth]{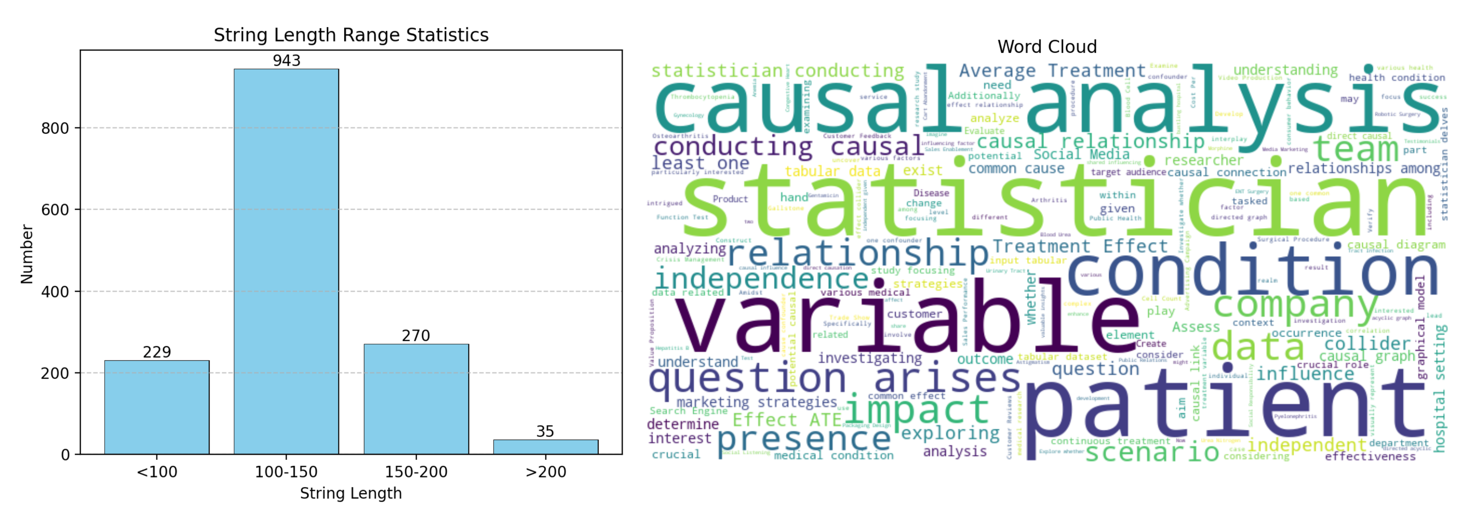}
    \caption{Diverse question length distribution and word cloud to ensure question diversity.}
    \label{fig:tongji}
\end{figure}

\begin{table}[h]
\centering
  \caption{The statistics about the dataset.}\label{biao:tongji}
  \begin{tabular}{cc}
    \toprule
     Attribute & \#number \\
    \midrule
    Question Template & 81\\
    Variables Pool Size & 311\\
    \bottomrule
  \end{tabular}
\end{table}
As shown in Table \ref{biao:tongji}, the CausalTQA is generated from 81 question templates, and data is sampled from a variable pool consisting of 311 variables to ensure the diversity of questions. Regarding the distribution of question lengths, the majority of questions in our synthetic dataset range from 100 to 150 words, while a smaller portion exceeds 200 words, thereby contributing to the diversity of the evaluation set, as shown in Figure \ref{fig:tongji}.

\subsection{Causal Problem Test Result on CausalTQA}\label{Causal Problem Test Result}
\begin{figure}
    \centering
    \includegraphics[width=1\linewidth]{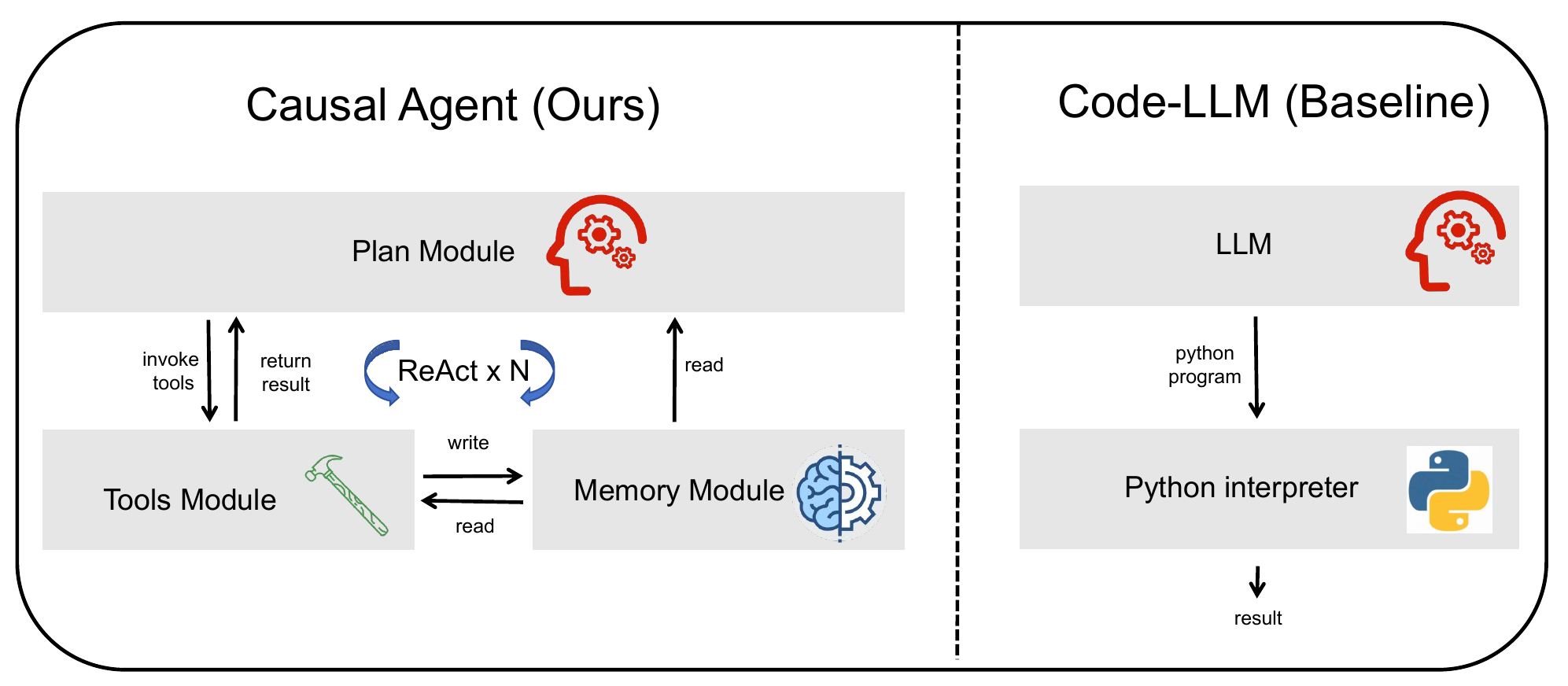}
    \caption{The comparison between our \uline{C}ausal \uline{A}gent(CA) and baseline(code-LLM).}
    \label{fig:duibi}
\end{figure}

The causal agent was constructed by GPT-3.5. Specifically, we use the gpt-3.5-turbo model default and set the temperature parameter as 0.5 when the causal agent reasoned. 
To constrain the output of the LLM and facilitate comparison with the actual ground truth during testing, we guide the model to output "yes", "no", and "uncertain" according to the question at the variable level and the edge level.

For causal graph-level questions, the agent would generate a causal graph during the reasoning process, and we would directly assess whether the causal graph was correctly generated. For causal effect-level questions, we considered whether the agent's calculation of the average causal effect was accurate. 



To compare with the original LLM without specially designed causal tools and verify the effectiveness of the causal agent framework, we also designed a baseline (referred to as code-LLM) comparison experiment. 
In code-LLM, we let code-LLM write code for causal problems and run the code once in the programming environment to realize the processing of tabular data. For code-LLM to use input and output correctly during reasoning, we write sample code for the prompt to process the format of the input and output so that code-LLM can follow it.
\begin{table*}[h!]
\centering
\caption{The test results of the causal agent on variable-level problems (IT, CIT, MCIT), causal graph level (TOTAL, PARTIAL), and causal effect level.}
\label{tab:combined_results}
\resizebox{1\textwidth}{!}{ 
\begin{tabular}{c|cc|cc|cc|cc|cc|cc}
\toprule
\multirow{2}{*}{\#node} & 
\multicolumn{2}{c}{\textbf{IT}} & 
\multicolumn{2}{c}{\textbf{CIT}} & 
\multicolumn{2}{c}{\textbf{MCIT}} & 
\multicolumn{2}{c}{\textbf{TOTAL}} & 
\multicolumn{2}{c}{\textbf{PARTIAL}} & 
\multicolumn{2}{c}{\textbf{Causal Effect Level}} \\ 
\cmidrule(lr){2-3} \cmidrule(lr){4-5} \cmidrule(lr){6-7} \cmidrule(lr){8-9} \cmidrule(lr){10-11} \cmidrule(lr){12-13} 
& CA (Ours) & Code-LLM & CA (Ours) & Code-LLM & CA (Ours) & Code-LLM & CA (Ours) & Code-LLM & CA (Ours) & Code-LLM & CA (Ours) & Code-LLM \\

\midrule
3  & 95.0 & 50.0 & 100.0 & 18.2 & 100.0 & 22.7 & 90.9 & 50.0 & -- & -- & 100.0 & 30.0 \\
4  & 95.0 & 35.0 & 100.0 & 31.8 & 100.0 & 9.1  & 63.6 & 54.5 & 95.5 & 40.9 & 100.0 & 35.0 \\
5  & 95.0 & 30.0 & 100.0 & 40.9 & 100.0 & 31.8 & 100.0 & 59.1 & 77.3 & 22.7 & 100.0 & 35.0 \\
6  & 95.0 & 15.0 & 100.0 & 36.4 & 100.0 & 40.9 & 90.9 & 54.5 & 95.5 & 36.4 & 100.0 & 40.0 \\
7  & 100.0 & 35.0 & 95.4 & 22.7 & 95.5 & 22.7 & 77.3 & 72.7 & 95.5 & 27.3 & 85.0 & 50.0 \\
8  & 95.7 & 43.5 & 100.0 & 21.1 & 100.0 & 36.4 & 86.4 & 45.5 & 86.4 & 36.4 & 100.0 & 35.0 \\
9  & 95.0 & 30.0 & 100.0 & 40.9 & 100.0 & 50.0 & 63.6 & 68.2 & 95.5 & 36.4 & 100.0 & 25.0 \\
10 & 90.0 & 55.0 & 100.0 & 22.7 & 100.0 & 59.1 & 81.8 & 68.2 & 95.5 & 27.3 & 100.0 & 35.0 \\

\midrule
Avg. & 95.1 & 36.8 & 99.4 & 29.4 & 99.4 & 34.1 & 81.8 & 59.1 & 91.6 & 32.5 & 98.1 & 35.0 \\ 
\bottomrule
\end{tabular}
}
\end{table*}
\subsubsection{The result of the variable level}



At the variable level, our results, as shown in the Table \ref{tab:combined_results}. We use IT to represent the independent test, CIT to represent the conditional independence test with one variable as a conditional variable, and MCIT to represent a conditional independence test beyond one variable as conditional variables. The causal agent achieved over 95\% accuracy across the three sub-questions of the variable level. Notably, in the conditional independence test, the agent correctly utilized the tools and reached the correct conclusions on almost all questions, achieving a 99.4\% in one conditional independence test and 99.4\% in a multi-conditional independence test, indicating our causal agent performed very well in this level.

We conducted a relevant analysis of the failure samples of the causal agent. The error of the causal agent may be caused by the  (1) \textbf{parameter passing error of the tool call.} For example, the question required an independence test for "Nephrology" and "General Surgery", not a conditional independence test. However, the causal agent mistakenly understood it as a conditional independence test, which resulted in the causal agent setting the condition parameter to other variables that appeared in the question when calling the variable-level tool. In addition, due to the misunderstanding of tools, causal agents may (2)\textbf{ misuse or use too many tools when making inferences}, resulting in a long logic chain, misleading the causal agent's reasoning, and causing the final result to be wrong. 

Compared with code-LLM, the accuracy improvement of our model at the variable level is significant. The code-LLM is not equipped with causal tools but directly lets LLM write Python code to process the tabular data. It is reasonable that code-LLM performs even worse than random because, in our experiments, code-LLM struggles to align tabular data with natural language and often makes basic input errors, even when we use in-context learning to guide the code-LLM to process tabular input through an example.

Specifically, code-LLM errors are mainly caused by two errors. (1)\textbf{ Context consistency error.} When code-LLM is writing code, it is possible to use parameter names incorrectly when passing function parameters. For example, the name "A B" incorrectly passes as "A\_B" when passing parameters. Morever, undefined and referenced library functions are used, such as using the norm function without defining it. Even if we prompt LLM to directly use the loaded data frame instead of reading and writing from the disk when the file is not imported or loaded, LLM will still reload the file incorrectly, causing IO problems. (2) \textbf{Code logic errors and grammatical errors.} Even if the consistency of the code context is maintained, the code itself can produce logical errors and syntax errors. For example, mismatching parentheses in the generated code may cause syntax errors, etc. Moreover, mathematics ability is currently difficult for LLM to master. LLM will produce mathematical errors when understanding statistics-related formulas and meanings.

\subsubsection{The result of the edge level}
\begin{table}[h]
\caption{The test results of the causal agent on edge-level problems.\label{edge_level}}
\centering
\begin{tabular}{cccc}
\toprule
\textbf{\#node}	& \textbf{CAUSE} & \textbf{COL}	& \textbf{CONF}\\
\midrule
3		& 95.0	& 100.0	& 95.5	\\
4		& 95.0	& 95.0	& 100.0	\\
5		& 90.0	& 95.0	& 100.0\\
6		& 80.0	& 95.7	& 100.0	\\
7		& 88.9	& 100.0	& 95.5	\\
8		& 83.3	& 94.4	& 86.4	\\
9		& 88.9	& 100.0	& 95.5\\
10		& 94.4	& 100.0	& 86.4	\\
\midrule
average		& 89.5  & 97.4	& 94.6	\\
\bottomrule
\end{tabular}

\end{table}

At the edge level, we tested the agent's accuracy in judging direct causal relationships (represented by CAUSE), confounding factors (represented by CONF), and colliders (represented by COL), with the results shown in Table \ref{edge_level}. Specifically, the agent achieved 89.5\% accuracy in judging direct cause relationships on average, 97.4\% accuracy in judging colliders on average, and 94.6\% accuracy in judging confounders.

Same as variable level, the failure case of the causal agent at the edge level mainly comes from \textbf{parameter passing error of the tool call.}

It should be noted that compared with our model, code-LLM is almost ineffective at this level of problem. Because code-LLM without causal tools does not have enough knowledge of the causal field and has difficulty understanding how to analyze and solve causal problems through code at this level. This process, which requires long-chain reasoning in the causal domain, is difficult for code-LLM, and the causal agent achieves better performance through modular encapsulation of causal tools.

\subsubsection{The result of the causal graph level}





At the causal graph level, we tested the agent's ability to generate a causal graph containing all variables (represented by TOTAL) and a partial causal graph (represented by PARTIAL) containing some variables. The specific results, as shown in Table \ref{tab:combined_results}, were  an 81.8\% accuracy rate for generating a causal graph with all nodes, and a 91.6\% accuracy rate for generating a partial causal graph composed of some nodes.

To constrain the causal graph output by code-LLM to be comparable, we restrict code-LLM to use the PC algorithm library of causallearn package through prompt and use ICL to provide a simple code example to inspire code-LLM. Through experiments, our causal agent is significantly better than code-LLM in generating causal graphs. In this task, code-LLM shows that there are still great limitations when generating causal graphs.

Analyzing the errors of causal agent and code-LLM, the main cause is insufficient understanding of the problem, which leads to errors in parameter transfer and tools' usage. When generating a complete causal graph,  the method of generating a partial causal graph is chosen mistakenly sometimes. This causes the parameters to be passed out of order, thus differing from the real causal graph. Regarding the error case of code-LLM, on the one hand, it is because, even if we provide the correct IO sample code, it still cannot follow the instructions and output the correct results. On the other hand, it is due to parameter errors. Similar to the previous section, the parameter names are wrong or the parameter order is confusing. 

\subsubsection{The result of the causal effects level}
At the level of causal effects, the causal agent answered 157 out of 160 questions correctly, achieving an accuracy of 98.1\%, as shown in Table \ref{tab:combined_results}. These examples cover the two fields of marketing and medicine and cover 3-10 nodes.

Three failure cases of the causal agent at this level are due to accidental parameter passing errors when invoking causal tools. Compared with severe hallucinations in code-LLM, causal agent effectively improves the reliability of model use through tools' modularity. While code-LLM, even though we provide examples for input and output as well as a causal effect function use case of EconML by ICL, it is still difficult for code-LLM to follow the instructions well and output the correct code, resulting in a decline in results.

\subsubsection{Causal agent cost calculation}\label{wentikaixiao}

We conduct overhead analysis on questions at the variable level, edge level, and causal graph level. In the experiments on CausalTQA, we set the maximum iteration count to 15. We randomly sampled 100 data points for testing. In addition, we categorize the questions into different types to examine whether the LLM’s computational cost and number of iterations vary across categories. The results are shown in Table \ref{tab:Statistics}.

\begin{table}[h!]
\centering
\caption{Overall Statistics and Iterations for Different Task Levels}
\label{tab:Statistics}
\begin{tabular}{lc}
\toprule
\textbf{Metric} & \textbf{Value} \\
\midrule
\multicolumn{2}{l}{\textit{Overall Averages}} \\
\quad Avg. Runtime & 3.37s \\
\quad Avg. Iterations & 2.51 \\
\quad Avg. Tokens & 5308 \\
\midrule
\multicolumn{2}{l}{\textit{Iterations by Task Level}} \\
\quad Variable Level & 2.35 \\
\quad Edge Level & 3.00 \\
\quad CG Level & 2.11 \\
\bottomrule
\end{tabular}
\end{table}

As the task becomes more complex, the average number of iterations also rises. Thus, complex tasks require more iterations to analyze and obtain the answer.

The average iteration count is 2.5. In practical applications, there is no need to dynamically design constraints on the maximum number of iterations, as this level of iteration is tolerable and does not bring too much additional cost and impact, which proves the stability of our framework.

\subsubsection{The impact of question domain types and answer types on the results.}
\begin{table}[h]
\caption{The test results after stratifying the answers and question domains are shown. Blue represents the medical domain, and red represents the market domain.
\label{table_results_after}}
\begin{tabular}{ccccccc}
\toprule
\textbf{answer}	& \textbf{IT} & \textbf{CIT}	& \textbf{MCIT} & \textbf{CAUSE}& \textbf{CONF} & \textbf{COL}\\
\midrule
\multirow{2}{*}{yes}		& \textcolor{red}{94.2}	& \textcolor{red}{100.0}	& \textcolor{red}{100.0} & \textcolor{red}{72.7}	& \textcolor{red}{90.9} & \textcolor{red}{87.5}	\\
		& \textcolor{blue}{97.7}	& \textcolor{blue}{97.6}	& \textcolor{blue}{100.0}	& \textcolor{blue}{92.9}& \textcolor{blue}{60.0}	& \textcolor{blue}{84.6} \\

\multirow{2}{*}{no}		& \textcolor{red}{96.4}	& \textcolor{red}{100.0}	& \textcolor{red}{100.0} & \textcolor{red}{87.3}	& \textcolor{red}{100.0} & \textcolor{red}{100.0}	\\
		& \textcolor{blue}{92.3}	& \textcolor{blue}{100.0}	& \textcolor{blue}{97.7}	& \textcolor{blue}{91.2}& \textcolor{blue}{98.5}	& \textcolor{blue}{98.5} \\

\multirow{2}{*}{uncertain}		& - 	& - & - & \textcolor{red}{100.0}	& \textcolor{red}{72.7} & \textcolor{red}{100.0}	\\
	& -	& -	& -	& \textcolor{blue}{100.0}& \textcolor{blue}{75.0}	& \textcolor{blue}{100.0} \\
\midrule
\multirow{2}{*}{average}		& \textcolor{red}{95.0}	& \textcolor{red}{100.0}	& \textcolor{red}{100.0} & \textcolor{red}{86.1}	& \textcolor{red}{94.5} & \textcolor{red}{98.6}	\\
		& \textcolor{blue}{95.2}	& \textcolor{blue}{98.8}	& \textcolor{blue}{98.8}	& \textcolor{blue}{92.5}& \textcolor{blue}{93.8}	& \textcolor{blue}{96.4} \\
\bottomrule
\end{tabular}
\end{table}



Additionally, we examine the correlation and impact between the ground truth of the problem and the answer's accuracy rate and explore how the types of domains involved in causal questions affect correctness. We conducted a stratified exploration based on the domains involved in the problems in our test set, which are the medical domain and market domain. Through stratification, we can see the impact of the problem domain and the answer on the results, as shown in Table \ref{table_results_after}. Different problem domains lead to different complexities and different descriptions of the problems, which affects the use of causal tools and the answers to the problems. But the two domains' correct rate differences are slight. Moreover, there are differences in the agent's accuracy under different answers to the questions. This suggests that the agent may exhibit output bias toward certain types of answers during the question-answering process. How to correct such bias will be left as a direction for future work.

\subsection{Discussion on Multi-Agent Systems and Different LLMs}\label{Discussion on Multi-Agent Systems and Different LLMs}

Causal agent focus on the tool interaction flow in the specific domain of causality, leveraging causal graphs as memory and intermediaries to link causal problem answers at different levels with various causal tools. This is a capability that current tool frameworks lack. In our causal agent, we not only call existing causal tools libraries like causal-learn and EconML, but our contribution also lies in the specialized interaction design and some manual alignment. 

To more comprehensively explore the impact of different frameworks on the ability of causal reasoning of tabular data, we have opted to use the RestGPT\cite{song2023restgpt} frameworks to encapsulate and test the APIs, which is a mainstream baseline model in the new framework for tool invocation research. Based on the tools used in our work, we constructed an agent and randomly selected 100 data points from CausalTQA for testing. The results are as shown in Table \ref{tab:restgpt_results}.
\begin{table*}[h]
\centering
\caption{RestGPT Results on 100 Data Points Random Selected from CausalTQA.}
\label{tab:restgpt_results}
\begin{tabular}{lcccccccc}
\toprule
\textbf{Model} & \textbf{IT} & \textbf{CIT} & \textbf{MCIT} & \textbf{CAUSE} & \textbf{COL} & \textbf{CONF} & \textbf{TOTAL} & \textbf{PARTIAL} \\
\midrule
RestGPT & 0.00 & 0.21 & 0.00 & 0.20 & 0.50 & 0.27 & 1.00 & 0.53 \\
\bottomrule
\end{tabular}
\end{table*}
Please note that the tools used are not entirely from existing tools, but include those we manually adapted and aligned with causal graphs in this work. In other words, causal graphs still serve as intermediaries for the short-term memory of tools and information exchange between tools. If we only use functions from the existing causal-learn library and the EconML library, the agent will be unable to continue with in-depth analysis of the causal graph, such as edge level, which is because the current library functions lack interfaces that can directly implement these functions.

Our analysis reveals that the complexity of the framework, along with the lack of causal logic and knowledge, actually makes it more difficult for LLMs to solve causal problems in interactions between multiple agents. The model lacks stability and is prone to logical confusion. We are unable to seamlessly integrate existing causal libraries into a powerful tool invocation framework, and the gap here is worth further investigation.

To explore the impact of different models on causal agents when they are the core of decision-making, we use the ChatGLM-4-air and ChatGLM-4-Plus models to conduct experimental analysis on market issues. The detailed results are shown in the Table \ref{tab:chatglm4air_results} and Table \ref{tab:ChatGLM-4-plus}.

As the foundational capabilities of the model we use increase, its accuracy on causal questions at different levels has significantly improved. This demonstrates the effectiveness of the framework, which can continue to improve as the foundational capabilities of LLMs expand. It validates the rationale of leveraging LLMs to empower automated causal reasoning.

\begin{table*}[h]
\centering
\caption{ChatGLM-4-air Results on CausalTQA.}
\label{tab:chatglm4air_results}
\begin{tabular}{c|cccccccc}
\toprule
\textbf{\#node} & \textbf{IT} & \textbf{CIT} & \textbf{MCIT} & \textbf{CAUSE} & \textbf{COL} & \textbf{CONF} & \textbf{TOTAL} & \textbf{PARTIAL} \\
\midrule
3      & 1.00 & 1.00  & 1.00 & 0.89  & 0.89 & 0.46  & 1.00  & 1.00    \\
4      & 0.90 & 0.91 & 1.00 & 0.78  & 0.67 & 0.46  & 0.82  & 1.00    \\
5      & 1.00 & 1.00 & 1.00 & 0.67  & 0.89 & 0.09 & 0.91  & 1.00    \\
6      & 0.90 & 1.00 & 1.00 & 0.56  & 0.78 & 0.28  & 1.00  & 1.00    \\
7      & 1.00 & 1.00 & 1.00 & 0.78  & 0.89 & 0.36  & 0.91  & 1.00    \\
8      & 1.00 & 0.91 & 1.00 & 0.56  & 0.89 & 0.27  & 1.00  & 1.00    \\
9      & 1.00 & 1.00 & 1.00 & 0.78  & 0.89 & 0.27  & 0.82  & 1.00    \\
10     & 0.70 & 1.00 & 1.00 & 0.33  & 1.00 & 0.45  & 0.91  & 1.00    \\
\midrule
Avg    & 0.94 & 0.98 & 1.00 & 0.67  & 0.86 & 0.33  & 0.92  & 1.00    \\
\bottomrule

\end{tabular}
\end{table*}

\begin{table*}[h]
\centering
\caption{ChatGLM-4-plus Results on CausalTQA.}
\label{tab:ChatGLM-4-plus}
\begin{tabular}{c|cccccccc}
\toprule
\textbf{\#node} & \textbf{IT} & \textbf{CIT} & \textbf{MCIT} & \textbf{CAUSE} & \textbf{COL} & \textbf{CONF} & \textbf{TOTAL} & \textbf{PARTIAL} \\
\midrule
3      & 0.93 & 1.00 & 1.00 & 1.00  & 1.00 & 1.00  & 1.00  & 1.00    \\
4      & 0.70 & 1.00 & 1.00 & 1.00  & 1.00 & 1.00  & 0.73  & 1.00    \\
5      & 0.90 & 1.00 & 1.00 & 1.00  & 1.00 & 1.00  & 1.00  & 1.00    \\
6      & 0.90 & 1.00 & 1.00 & 1.00  & 1.00 & 1.00  & 1.00  & 1.00    \\
7      & 0.90 & 1.00 & 1.00 & 1.00  & 1.00 & 1.00  & 1.00  & 1.00    \\
8      & 0.80 & 1.00 & 1.00 & 1.00  & 1.00 & 1.00  & 0.91  & 1.00    \\
9      & 0.80 & 1.00 & 1.00 & 1.00  & 1.00 & 0.91  & 0.91  & 1.00    \\
10     & 0.90 & 1.00 & 1.00 & 1.00  & 1.00 & 1.00  & 1.00  & 1.00    \\
\midrule
Average & 0.85 & 1.00 & 1.00 & 1.00  & 1.00 & 0.99  & 0.94  & 1.00    \\
\bottomrule
\end{tabular}
\end{table*}

\subsection{Testing on real-world tabular data}\label{QRDATA verify}
We expanded experiments to include evaluations on publicly available datasets from a subset of QRData for causal inference on tabular data, covering a range of diverse problems, and compared with a variety of mainstream baseline models. As shown in Table \ref{tab:QRData}. Our model demonstrates general robustness in real-world data scenarios. It is worth noting that although the base model we used is less powerful than GPT-4, it still outperforms the SOTA results. If we use the GLM-4-plus model, which is comparable to GPT-4, our model is 6\% higher than the SOTA.
\begin{table}[h]
\centering
\caption{Performance comparison of causal agent and other baselines on QRData data. The notation "$^{\dag}$" means results from \cite{liu2024llmscapabledatabasedstatistical}}\label{tab:QRData}
\begin{tabular}{@{}llc@{}}
\toprule
\textbf{Model} & \textbf{Size} & \textbf{Result} \\
\midrule
\textit{Random$^{\dag}$} & - & 27.2 \\
\midrule
\rowcolor{gray!10}
\multicolumn{3}{@{}l}{\textit{Table Question Answering$^{\dag}$}} \\
\hspace{1em}TableLlama\cite{zhang2023tablellama} & 7B & 12.6 \\
\rowcolor{gray!10}
\multicolumn{3}{@{}l}{\textit{Chain of Thought Prompting$^{\dag}$}} \\
\hspace{1em}Llama-2-chat\cite{touvron2023llama} & 7B &  23.0 \\
\hspace{1em}AgentLM\cite{zeng2023agenttuning} & 7B & 27.1 \\
\hspace{1em}WizardMath\cite{luo2023wizardmath} & 7B & 28.6 \\
\hspace{1em}CodeLlama-instruct\cite{roziere2023code} & 7B & 21.9 \\
\hspace{1em}Deepseek-coder-instruct\cite{guo2024deepseek} & 6.7B & 20.4 \\
\hspace{1em}Gemini-Pro\cite{team2023gemini} & - & 35.3 \\
\hspace{1em}GPT-4\cite{achiam2023gpt} & - & 42.8 \\
\midrule
\rowcolor{gray!10}
\multicolumn{3}{@{}l}{\textit{Program of Thoughts Prompting$^{\dag}$}} \\
\hspace{1em}Llama-2-chat & 7B & 1.5 \\
\hspace{1em}AgentLM & 7B & 0.4 \\
\hspace{1em}WizardMath & 7B &  8.9 \\
\hspace{1em}CodeLlama-instruct & 7B &  16.0 \\
\hspace{1em}Deepseek-coder-instruct & 6.7B &  32.3 \\
\hspace{1em}Gemini-Pro & - & 20.1 \\
\hspace{1em}GPT-4 & - &  36.8 \\
\midrule
\rowcolor{gray!10}
\multicolumn{3}{@{}l}{\textit{ReAct-style Prompting$^{\dag}$}} \\
\hspace{1em}Llama-2-chat & 7B & 15.2 \\
\hspace{1em}AgentLM & 7B & 13.4 \\
\hspace{1em}WizardMath & 7B &  18.2 \\
\hspace{1em}CodeLlama-instruct & 7B &  15.6 \\
\hspace{1em}Deepseek-coder-instruct & 6.7B &  21.2 \\
\hspace{1em}Gemini-Pro & - & 37.5 \\
\hspace{1em}GPT-4 & - & 51.3 \\
\midrule

\rowcolor{gray!10}
\multicolumn{3}{@{}l}{\textit{Code Interpreter Assistants$^{\dag}$}} \\
\hspace{1em}GPT-3.5 Turbo & - &31.2 \\
\hspace{1em}GPT-4 & - & 46.8 \\
\midrule

\rowcolor{red!20}
\multicolumn{3}{@{}l}{\textit{\textbf{Causal Agent (Ours)}}} \\
\hspace{1em}Causal Agent (GPT-3.5)  & - & \underline{52.4}(+1.1\%) \\
\hspace{1em}Causal Agent (GLM-4-Plus)  & - & \textbf{57.3} (+6.0\%) \\
\midrule

Human$^{\dag}$ & - & 68.8 \\
\bottomrule
\end{tabular}%
\end{table}



\section{RELATED WORK}
\textit{Causality}. As a tool for data analysis, causality aims to accurately identify and quantify the actual effects of specific factors (causes) on outcome variables (effects) in a complex system environment. It is everywhere in our daily lives, such as statistics\cite{chilvers1988journal,heckman2022causality,berzuini2012causality}, economics, computer science\cite{10.1145/3652610,10.1145/3714430}, epidemiology\cite{vlontzos2022review,gottlieb2002relational,10.1145/3604809} and psychology\cite{child1994causality}. Pearl's "Ladder of Causality" theory divides causality into three progressive levels\cite{pearl2018book}: association, intervention, and counterfactual. The core purpose of studying causality is to reveal the true causal chain between things and to abandon those confusing pseudo-causal relationships. Causal field problems can be briefly divided into two broad directions: causal discovery and causal inference. Causal discovery is based on directed acyclic graphs and Bayesian models, focusing on obtaining causal relationships from observation data, and methods can be divided into constraint-based methods\cite{tu2019causal,spirtes2013causal}, such as IC, PC, FCI, and function-based methods\cite{shimizu2006linear,shimizu2011directlingam,hoyer2008nonlinear} such as LiNGAM and ANM, and hybrid methods\cite{chickering2002optimal} to combine the advantages of the above two methods. Common frameworks for causal inference are the structural causal model and the potential outcome framework\cite{rubin1974estimating}.  Researchers use the structural causal model and Rubin Causal Model to model the interaction between variables and calculate causal effect estimates such as average treatment effect (ATE) and conditional treatment effect (CATE). 

\textit{LLM-based Agent}. Autonomous agents have long been considered a promising approach to achieving artificial general intelligence (AGI), which accomplishes tasks through autonomous planning and action\cite{9169794,10.1145/3744746}. Human-like intelligence has shown great potential\cite{achiam2023gpt,touvron2023llama,10.1145/3793671} and there has been a large amount of research using LLM as the decision-making and reasoning center of agents\cite{shen2024hugginggpt,qin2023toolllm,schick2024toolformer}, achieving great success in natural sciences\cite{boiko2023emergent,swan2023math}, engineering sciences\cite{mehta2023improving,wu2024chateda,qian2023communicative,10.1145/3714474}, and human simulation\cite{kovavc2023socialai,fischer2023reflective}. The LLM agent is composed of four parts, namely the profile module, action module, plan module, and memory module. The profile module assigns an imaginary role to the agent, such as a teacher or poet. The planning module helps the agent use thinking chains to break down tasks and use different search methods to obtain solutions in the problem space, such as  CoT\cite{wei2022chain}, ToT\cite{yao2024tree}, AoT\cite{sel2023algorithm}, Reflexion\cite{shinn2024reflexion}, etc. The memory module is subdivided into two categories: short-term memory and long-term memory. The action module is the key for the intelligent agent to take specific actions in the physical or virtual environment.

\textit{Combining LLM and Causality}. Since the advent of LLM, some researchers have evaluated and analyzed the causality ability of LLM \cite{wu2024causality,10347447,10361523,10304625,10028777,9052492}. \cite{jin2023large} introduced a new task CORR2CAUSE, which can infer causal relationships from correlations, to evaluate the causal inference ability of large models. From the experimental results, it is generally believed that the LLM with a higher version or better reasoning ability does not show positive correlation results in the causal inference task, and the performance of the LLM in the causal inference task is akin to random. \cite{jin2024cladder} further investigated whether large language models can reason about causality and proposed a new NLP task: causal inference in natural language. They built a large dataset, CLADDER, with 10K samples. However, the problem is described in natural language, making it incapable of handling tabular data. \cite{gao2023chatgpt} presents a comprehensive evaluation of ChatGPT's causal reasoning capabilities. They found that ChatGPT is not a good causal reasoner, but is good at causal explanation and that ChatGPT has a serious problem of causal hallucinations, which is further exacerbated by In-Context Learning (ICL) and chain of thought techniques. \cite{zečević2023causal} argues that large language models cannot be causal and define a new subgroup of structure causal models, called meta-SCMs. Their empirical analysis provides favorable evidence that the current LLM is even weak "causal parrots". \cite{kiciman2023causal} found LLM can achieve competitive performance in determining pairwise causality, with an accuracy of up to 97$\%$, but their performance varies depending on the quality of cue word engineering. \cite{liu2024llmscapabledatabasedstatistical} proposed QRData, which is the first attempt to model causal and statistical issues from the perspective of LLMs handling tabular data. However, his focus is on evaluation, and he does not propose ways to enhance the ability of LLMs to address causal issues in tabular data.

\textit{The difference of this work.} (a) Our work is the first to model table-based causal question answering from the perspective of LLMs across four levels and regarding LLMs as agents for automating causal reasoning. (b) We focus on the interaction flow between tools and propose a systematic causal agent. (c) We propose CausalTQA, which has about 1.4K questions covering four levels of causal questions.

\section{CONCLUSION}

In this work, we harnessed LLM to construct a causal agent by invoking causal tools, modeling causal problems across four levels in the causal domain, and endowing the large model with causal capabilities at four levels. To test the ability to deal with the causal question answer on tabular data, we propose the CausalTQA benchmark, which has 1.4K problems at four causal question levels. The experiments of the causal agent on CausalTQA showed that it performed particularly well. Through comparisons of multi-agent frameworks, error case analyses, the use of different LLM-based decision cores, and cost evaluations, we systematically analyzed the advantages of the causal agent and its potential future research directions. Through comparison with real-world data from QRData, the causal agent shows strong generalization ability. The causal agent framework provides a feasible solution for improving the causal ability of large models and automated causal reasoning in real-world scenarios.

\bibliographystyle{ACM-Reference-Format}
\bibliography{sample-base}

\appendix









\end{document}